\documentclass[journal]{IEEEtran}
\usepackage{xcolor}
\usepackage{soul}

\ifCLASSINFOpdf
   \usepackage[pdftex]{graphicx}
\else
\fi
\usepackage{amsmath}
\usepackage{amssymb}

\usepackage{stfloats}

\usepackage{algorithm}
\usepackage{algorithmic}
\begin{document}
\title{Memory-efficient training with streaming dimensionality reduction}

\author{Siyuan Huang,
		Brian D. Hoskins,
        Matthew W. Daniels,
        Mark D. Stiles,
        and Gina C. Adam% <-this % stops a space
\thanks{S. Huang and G. Adam are with the Department
of Electrical and Computer Engineering, George Washington University, Washington,
DC, 20037 USA \newline   e-mail: ginaadam@gwu.edu.}% <-this % stops a space
\thanks{B. Hoskins, M. Daniels and M. Stiles are with the Physical Measurement Laboratory, National Institute of Standards and Technology, Gaithersburg, MD, 20899 USA.}}% 
\maketitle

\begin{abstract}
The movement of large quantities of data during the training of a Deep Neural Network presents immense challenges for machine learning workloads. To minimize this overhead, especially on the movement and calculation of gradient information, we introduce streaming batch principal component analysis as an update algorithm. Streaming batch principal component analysis uses stochastic power iterations to generate a stochastic $k$-rank approximation of the network gradient. We demonstrate that the low rank updates produced by streaming batch principal component analysis can effectively train convolutional neural networks on a variety of common datasets, with performance comparable to standard mini batch gradient descent. These results can lead to both improvements in the design of application specific integrated circuits for deep learning and in the speed of synchronization of machine learning models trained with data parallelism.

\end{abstract}

\begin{IEEEkeywords}
Deep Learning, Gradient data decomposition, Streaming, Principal Component Analysis.
\end{IEEEkeywords}

% \fi
\IEEEpeerreviewmaketitle

\section{Introduction}
\IEEEPARstart{S}{tochastic} gradient descent (SGD) and minibatch gradient descent (MBGD) are two common means of training a neural network, in which subsets of the input data are used to compute approximations to the gradient of the loss function more frequently but less accurately than computing it for the entire training set. Pipelining (starting the processing of each input before the previous is completed) the input in each batch gives speed-ups that improve with batch size giving a trade off between the wall-clock training speed with the accuracy gains achieved from stochastic search of the space \cite{shallue2019measuring}. 
% ^ It's not clear to me that this is a fair characterization of the "tradeoff" between SGD and MBGD --mwd
From batch to batch, however, significant compute and memory overheads are required to calculate and store the weight updates. More advanced training methods, such as momentum,  require more memory  to train the network than to store the network itself. Managing the movement of such large quantities of data is increasingly challenging and energy intensive, especially for large networks in data center environments \cite{mayer2020scalable,li2014communication,wen2017terngrad}. 

Taking advantage of these approaches to training requires collocating the training operations with the inference operations. This requirement is particularly true for approaches that greatly increase the efficiency of inference.  Without more efficient training, these advantages are minimized when the whole system is considered.  For example, the use of emerging hardware approaches for nonvolatile memory, such as crossbar arrays \cite{kataeva2019towards,burraccelerating,adam20163,adam2018challenges,chakrabarti2017multiply,prezioso2015training}, promises great efficiency gains for inference by neural networks. While such arrays are able to efficiently store the network itself, alternative forms of memory management are needed to store the training data and transfer it into the long term memory array. 

Streaming techniques can offer substantial memory reduction because they process each input data string and accumulate the results in a efficient way. Data decomposition methods, like Principal Component Analysis (PCA) and its reduced rank variants, also reduce storage and processing requirements by only computing the most important contributions to the gradient. Previously, we showed that a low rank approximation of MBGD can train networks with many of the same benefits of MBGD by combining streaming methods with  PCA to produce rank-1 approximations of batch updates \cite{hoskins2019streaming}. 

Here, we significantly expand on that initial proposal by developing an arbitrary rank approximation of MBGD called streaming batch principal component analysis (SBPCA). Instead of using deflation to expand the number of ranks, as proposed previously, we investigate a bi-iterative PCA algorithm based on $QR$-factorization. This method uses Gram-Schmidt orthonormalization or Householder reflections to ensure orthogonality between the ranks~\cite{hoskins2019streaming,yang1995extension,mitliagkas2013memory}.  Additionally, by investigating both the rank and data set dependence of our algorithm and applying the technique to much larger data sets, we are able to demonstrate accurate training and convergence across a broad range of potential use cases. 

Our main contributions are as follows:
\begin{itemize}
    \item We show our proposed streaming batch principal component analysis can recompose an accurate  gradient matrix with a few ranks, with 10 ranks often achieving near equivalent accuracy to minibatch gradient descent;
    \item We show that streaming batch principal component analysis is functionally correct, that is, it can track the correct updating direction and minimize the loss function; and
    \item We show that streaming batch principal component analysis converges to near equivalent accuracy with significant memory savings.
\end{itemize}

The remainder of the paper is organized as follows. Section II introduces background information related to our method. Section III describes the problem and our detailed algorithm. Section IV presents the evaluation of the algorithm on three real data sets. Section V concludes with a final discussion. 

\section{Related Work}

Many different approaches have been proposed to improve the efficiency of training neural networks.  In this section, we discuss several that motivate the present approach.

\subsection{Acceleration of Deep Neural Networks}
As the market demand for deep neural networks grows, both data centers and deep neural network frameworks are constantly being improved to accommodate the ever changing demands of  machine learning workloads. To fully exploit the hardware available within the data center, deep neural network frameworks will have to natively support methods for model and data parallelism. Ideally, this support would allow for a linear scaling of the model training speedup with the number of available computing resources. These hardware resources include traditional central processing units (CPUs), graphical/tensor processing units (GPUs/TPUs), as well as data communication buses such as peripheral connection interconnect express (PCIe) buses. 

The most straightforward method, data parallelism, simply duplicates the machine learning model across all of the available resources as efficiently as possible \cite{shallue2019measuring,mayer2020scalable}. These multiple instances can then be used to process batches of the data in parallel. In between processing the data, weight and gradient information is transmitted throughout the computing platform so that the duplicate models can be updated. The synchronization of the network models is a time consuming process due to the limited communication bandwidth available and the large size of the models. In some cases, communication of the gradient and weight information can be the most time consuming step during the training process, taking up more than 60~\% of the overall training time \cite{wang2019characterizing}. Even in hybrid approaches using model parallelism, where separate parts of the model can operate independently but synchronously or even asynchronously, transmission of gradient information and model synchronization are important bottlenecks. 

To deal with this problem, different data reduction approaches have been considered. The simplest, gradient quantization, is to reduce the precision of the gradient information transmitted \cite{mayer2020scalable}. Other approaches, such as gradient sparsification, transmit only the largest values of the gradient, or, only sparsely communicate the weight updates when they reach a threshold of significance \cite{mayer2020scalable}. Much effort has been invested in these approaches, enabling reductions in the volume of gradient data  transmission by several hundred times without reducing accuracy \cite{lin2017deep}. However, such approaches may have high local memory and computational overhead, requiring significant logic to be performed over the full set of estimated gradients and weights within the parallel model. 
%Efficient compression of the gradient information is therefore an important area of investigation. 

Principal component analysis provides an efficient method for compressing the gradient information, by extracting the most important components.  As such, it could significantly contribute to the effectiveness of parallelization of gradient updates.

\subsection{Challenges and Opportunities in Emerging Hardware}
The movement of data, weight matrices, and gradients is a major source of latency and resource utilization in deep neural networks. One approach to minimizing this movement is to build nonvolatile memories that can simultaneously store the weights and compute the multiply and accumulate operations commonly implemented in GPUs and TPUs~\cite{adam2018challenges,prezioso2015training}. Performing these operations in the nonvolatile memory itself eliminates the cycling of data in and out of the local GPU/TPU memories.  

The principal advantage afforded by this scheme is the amount of information that can be stored compared to other approaches. On one hand a TPU can store approximately 65 kilobytes of weight data within its primary systolic array, which is only a small fraction of an overall deep neural network model. On the other hand, a nonvolatile memory array based on flash memories, resistive random accesses memory (ReRAM), or magnetic tunnel junctions, could potentially store terabytes of model data at much greater density~\cite{jouppi2017datacenter,adam2018challenges}. Different methodologies for implementing the operation of these arrays include both analog and digital approaches, such as binary neural networks~\cite{burraccelerating,prezioso2015training,suri2013bio}. In either case, multiply and accumulate operations are encoded in charge and current summation, natively implemented via Kirchoff's law, allowing for massively parallel multiplication of the model data with the input activation or error. One such emerging computing architecture that potentially exploits this physics is the three dimensional system on a chip (also known as 3DSoC) architecture. The 3DSoC incorporates dense vertical ReRAM with both silicon and carbon-nanotube digital logic~\cite{bishop2019monolithic}.

The principal weaknesses of such approaches include the high programming energy of nonvolatile memories, which would favor sparse updates to the network, as well as the difficulty of storing and processing the immense quantities of training and gradient data. Whereas the model can be stored in the nonvolatile memory arrays, there is at present no compatible ultra dense short term memory array with which to store the training and gradient data, though potential approaches are under investigation~\cite{li2018capacitor}. Existing implementations of L1 and L2 caches could be used to temporarily store this data for transfer into the nonvolatile memory arrays, but efficient compression of the training and gradient data would be required. 

Batch PCA significantly reduces the number of hardware updates for each training data set. It allows for outer-product updates in which all of the components are updated at once in the most important directions.  It also compresses a number of input data into a single best average update further reducing the rate of hardware updates.

\subsection{Streaming Principal Component Analysis}

\begin{figure*}[!ht]
\centering
\includegraphics[trim=0in 9in 1in 0in,clip,width=\textwidth]{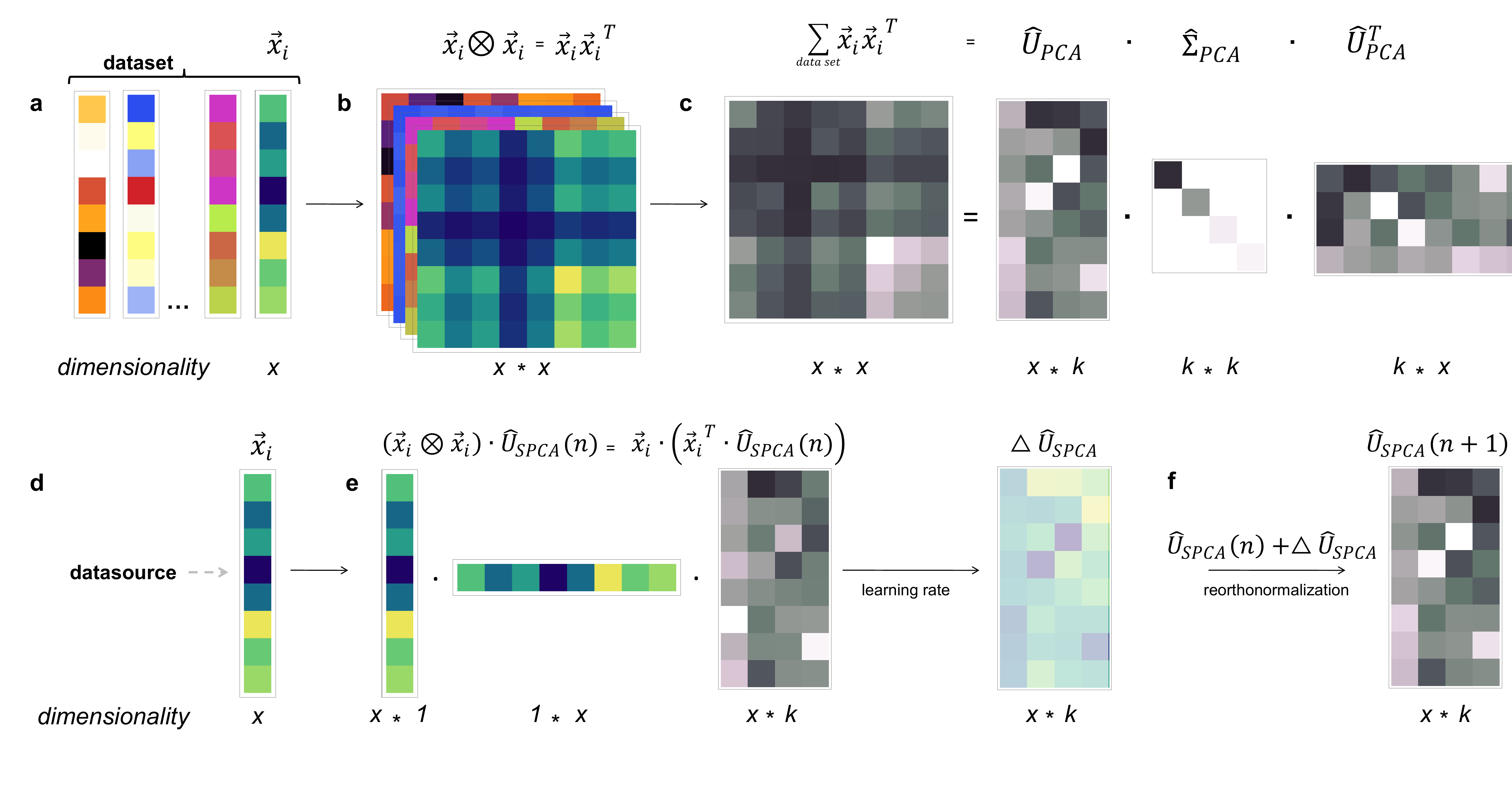}
% where an .eps filename suffix will be assumed under latex, 
% and a .pdf suffix will be assumed for pdflatex; or what has been declared
% via \DeclareGraphicsExtensions.
\caption{Traditional vs. Streaming PCA algorithms. (a) A dataset of examples $\vec{x}_i$ is composed into a covariance matrix by (b) taking and summing the outerproducts. This covariance matrix is (c) decomposed into its principal components $\hat{U}_{PCA}$ by taking the singular value decomposition. Instead of the full data set, streaming methods can use (d) individual examples to (e) calculate an update $\Delta\hat{U}_{SPCA}$ to an approximate set of principal components $\hat{U}_{SPCA}(n)$ and produce (f) an updated set $\hat{U}_{SPCA}(n+1)$. This sequential updating process can avoid the computation and memory intensive outerproduct operations.}
\label{FigStreaming}       
\end{figure*}

Memory-efficient compression of large matrices without explicitly calculating them has a robust history of theoretical investigation~\cite{oja1985stochastic,liu1989limited,guan2012online}. For one approach, streaming principal component analysis (SPCA), the objective is to extract the top $k$ principal components ($\hat{U}_{PCA}$) from a stream of incoming data  $\{\vec x_i\}$. Due to the high dimensionality of $\vec x_i$,  calculation of the covariance matrix $\sum_i\vec x_i\vec x_i^T$ and its singular value decomposition (SVD) would be costly, particularly in terms of memory, as the covariance matrix (as in Fig. \ref{FigStreaming}) might not even be able to fit within the available memory.
Inspired by neural biology, Oja proposed a linearized version of stochastic power iterations that can be used to extract one, or, with deflation, multiple principal components of a random data stream~\cite{oja1982simplified,oja1992principal}. More modern approaches, such as history PCA, rely instead on $QR$-factorization to ensure rank orthonormality~\cite{yang2018history,hardt2014noisy,li2016rivalry,mitliagkas2013memory}. Update and reorthonormalization of the $Q$ matrix allows the algorithm to span a greater fraction of the input data space, reducing the effective eigengap of the stochastic power iteration and improving the rate of convergence\cite{allen2017first,balcan2016improved}.

As we discuss below, our proposed approach does not map perfectly to streaming PCA; we are trying to extract the singular values of a general rectangular gradient matrix, rather than a square and symmetric covariance matrix. We combine streaming PCA methods with bi-iterative power iterations to perform stochastic approximation of the top singular vectors. Such an approach was first used by Strobach to perform subspace tracking \cite{strobach1997bi}. To the best of our knowledge, such an approach has not been used previously to extract the top $k$ principal components of a neural network gradient matrix during training.

\section{Method Details}

\subsection{Low Rank Gradient Descent}
In neural networks, finding a solution to a classification problem is, for a given network structure of weights and neurons, mapped to minimizing a scalar loss function, $\ell$, which correlates inputs with correct classifications. In classical gradient descent, we define the differential of the loss function $d \ell$ over all of the weights $\vec\Theta$ in a layer as
\begin{equation}
   d \ell = {\frac{\partial \ell}{\partial \vec\Theta}} \cdot d\vec\Theta.
\end{equation}
During training, we compute $\partial \ell/\partial \vec\Theta$ at every step. The negative of this gradient gives the direction of steepest descent. Choosing a step of $\Delta\vec\Theta= -\alpha (\partial \ell/\partial \vec\Theta)$ changes the loss function by  
\begin{equation}
    \Delta \ell = -\alpha \Vert \nabla_{\vec\Theta} \ell \Vert^2,\label{eq:delta-ell}
\end{equation}
where $\alpha$ is the learning rate.

Though we have presented the weights in vector form thus far, we could instead suppose that they are arrayed in a matrix $\hat\Theta$, as is typical in neural networks. In that case we recognize the norm taken in Eq.~\eqref{eq:delta-ell} as the Frobenius norm of $\nabla_{\hat\Theta}\ell$. It is well known that Eq.~\eqref{eq:delta-ell} can then be written as
\begin{equation}
    \Delta \ell = -\alpha \sum_j \sigma_j^2,
    \label{eq:frobenius-sum}
\end{equation}
where the $\sigma_j$ are the singular values of $\nabla_{\hat\Theta}\ell$ in its singular value decomposition. This leads us to the following observation. Suppose $\nabla_{\hat\Theta}\ell$ could be well-approximated by a truncated singular value decomposition using only the first $k$ ranks. We write this approximation as $\nabla_{\hat\Theta}^{(k)}\ell$. Then moving along $\nabla^{(k)}_{\hat\Theta}\ell$ instead of $\nabla_{\hat\Theta}\ell$ produces $\Delta\ell$ almost as large as the exact gradient descent, up to a small error on the order of $O(\sigma_{k+1}^2/\sigma_1^2)$. We conjecture that using the truncated decomposition to propose weight matrix updates may still allow one to minimize the loss function, but with a smaller memory footprint than would be needed to store the full-rank gradient matrix.

In practice, $\Delta\hat\Theta$ is typically not computed with the true gradient of $\ell$, but a stochastic approximation of the gradient over a relatively small batch of samples of size $B$, much less than the size of the full dataset. This balances the acceleration of training available from the pipelining of $B$ inferences at a time against the desired stochasticity of the sampled gradient. For our proposed method, the equivalent stochastic algorithm would generate an approximation of $\tilde\nabla^{(k,B)}_{\hat\Theta}\ell$ which is randomly sampled from a $B$-batch of the overall training set and is also of tunable truncation rank $k$. 

\subsection{Low Rank Stochastic Approximation}
\begin{figure*}[!ht]
\centering
\includegraphics[trim=0in 14.2in 2in 0in,clip,width=\textwidth]{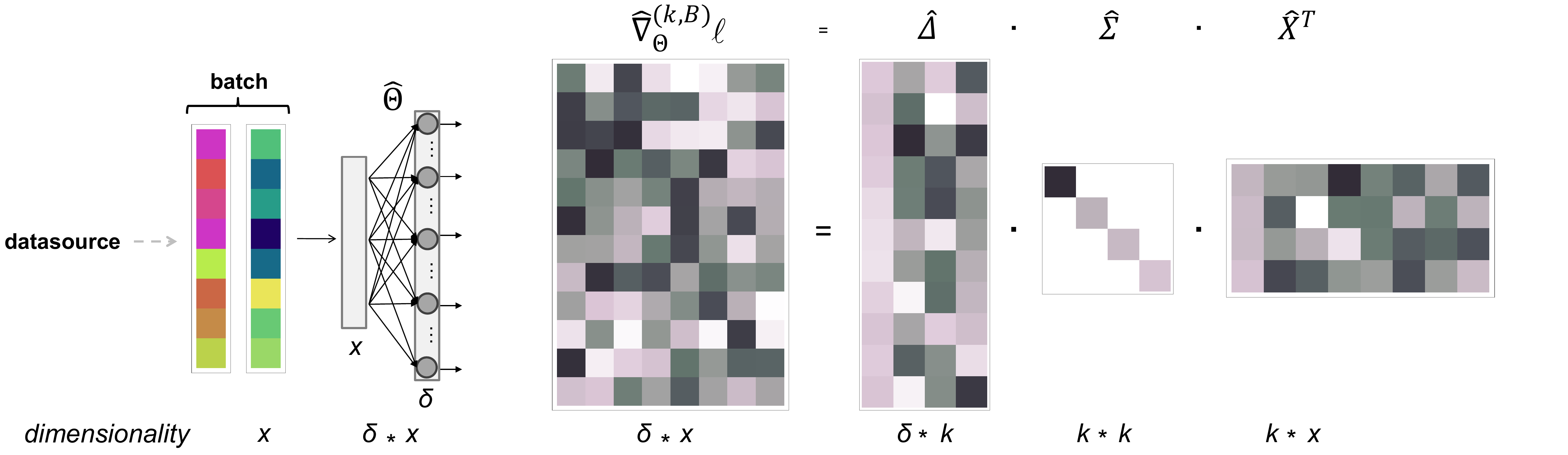}
% where an .eps filename suffix will be assumed under latex, 
% and a .pdf suffix will be assumed for pdflatex; or what has been declared
% via \DeclareGraphicsExtensions.
\caption{Breakdown of a full rank gradient matrix in a neural network layer into its low rank approximation. For a given set of weights $\hat{\Theta}$, a stochastic  rank $k$ approximation of over a batch $B$ can be defined as $\tilde\nabla_\Theta^{(k,B)}\ell$. This low rank approximation can be represented in decomposed forms with the left singular vectors matrix $\hat{\Delta}$, the singular value matrix $\hat{\Sigma}$, and the right singular vectors matrix $\hat{X}^T$. The decomposed form has less memory overhead than the full rank matrix for sufficiently few ranks.}
\label{FigSBPCA}
\end{figure*}

In order to generate a $k$-rank stochastic approximation of the matrix $\Delta\hat\Theta \in \mathbb{R}^{m\times n}$, we start from the truncated singular value decomposition for the approximate gradient
\begin{equation}
    \tilde\nabla_\Theta^{(k,B)}\ell = \hat\Delta\hat\Sigma \hat{X}^T ,
\end{equation}
where $\hat\Delta\in \mathbb{R}^{m \times k}$ is a matrix composed of the top $k$ left singular vectors (that is, singular backpropagated errors) and $\hat{X}\in \mathbb{R}^{n \times k}$ is a matrix of the top $k$ right singular vectors (that is, singular activations). $\hat\Sigma$ is a square $k\times k$ matrix with singular values $\vec\sigma$ on the diagonal and zeroes elsewhere. Fig. \ref{FigSBPCA} depicts the breakdown and dimensionality of this decomposition with respect to the weight matrix in the network layer. The memory cost of this decomposed form is $mk + k + nk = k(m + n + 1)$, which, for sufficiently small $k$, is significantly less than the memory overhead of the full $m\times n$ matrix. The challenge now is to determine the correct values of $\hat{X}$, $\hat\Delta$, and $\vec\sigma$ in a computationally efficient way, given a list of input activations $\vec{x}_i$ and backpropogated vectors $\vec\delta_i$ of a fully connected layer in a neural network. 

Performing the singular value decomposition using traditional methods will fail to deliver efficient memory and compute footprints. Suppose that for a particular batch of inputs $\{\vec x_i\}$ and errors $\{\vec \delta_i\}$ one constructs the update matrix $\Delta\hat\Theta = \sum_i \vec\delta_i\vec x_i^T$ as usual and then takes its full singular value decomposition. In addition to requiring the calculation [$O(Bmn)$ operations] and storage [$O(mn)$] of the full array, an additional $O(n^2m)$ operations (assuming $n<m$) are required to perform the singular value decomposition. 

Calculations of the truncated SVD requiring only the top $k$ ranks can be dramatically more efficient, with a complexity of $O(k^2m)$. Like Lanczos's method, these methods are based on the use of power iterations to extract the top eigenvalues and eigenvectors. If these power iterations are replaced with \emph{stochastic} power iterations---that is, if the gradient is replaced by a stochastic approximation from the input $\{\vec x_i\}$ and $\{\vec \delta_i\}$ values---then the computational complexity can be dramatically reduced, and in fact it is possible to avoid directly calculating or storing the batch gradient. Since the batch already gives a stochastic approximation of the gradient, it is plausible that further stochastic approximation may not yield significantly reduced performance, while reducing memory and computational overheads.

\subsection{Streaming Principal Component Analysis Batch Update}

The general approach for a multi-rank update of the $\hat{X}$ and $\hat\Delta$ matrices follows directly from conventionally used algorithms in streaming PCA, with the caveat that the target matrix is a general rectangular matrix rather than a square and symmetric covariance matrix. From iteration to iteration $\hat{X}$ can be updated as:
\begin{equation}
    \hat{X} \leftarrow (1-c)\hat{X} + c\hat{d} ,
\end{equation}
where $c$ is a convergence coefficient which gradually decays to zero.  The matrix $\hat{d}$ is an update to $\hat{X}$ based on block-size $b$ entries of new data such that
\begin{equation}
    \hat{d} = \frac{1}{b}{\hat{x}^T \hat\delta \hat\Delta\hat\Sigma^{-1}} ,
\end{equation}
where $\hat{x}\in \mathbb{R}^{b\times n}$ and $\hat\delta\in \mathbb{R}^{b\times m}$ are matrices constructed from $b$ sets of $\vec{x}$ and $\vec\delta$ respectively.
After updating the values of $\hat{X}$, the matrix can be reorthonormalized using $QR$-factorization, where
\begin{equation}
\hat{X} \leftarrow {Q}{R}(\hat{X})
% Consider, perhaps: (X, --) <- QR(X)
\end{equation}
defines the $QR$-factorization of $\hat{X}$ with the $R$ matrix discarded, and, to ensure the uniqueness of the factorization, the $R$ matrix is defined with non-negative entries on the diagonal. The matrix $\hat\Delta$ can be updated in a reciprocal fashion, making this algorithm fall within the  family of bi-iterative stochastic power iterations. The vector $\vec\sigma$ can be similarly updated with a block average of the data by 
\begin{equation}
\vec\sigma \leftarrow (1-c)\vec\sigma + c\sum_{\text{rows}} \frac{(\hat\delta_i \hat\Delta) \odot (\hat{x}_i \hat{X})}{ b} ,
\end{equation}
which performs inner products over the singular vectors with the incoming data streams to determine the singular values of the matrix.  The matrix operation evaluates as $\left(\sum_{\text{rows}} A \odot B\right)_j = \sum_i A_{ij} B_{ij}$. The rank-wise renormalization of the update $\hat{d}$ in Eq.~6 is critical as it ensures that the learning rate is consistent from epoch to epoch. As the accuracy of the network increases, the magnitude of the input activations would remain approximately constant, but the backpropogated errors' magnitudes will decline, reducing the effective learning rate during training if they are not rescaled. In classic streaming PCA approaches, this role is filled by adaptively choosing the learning rate. 

Since the $QR$-factorization is the most computationally intensive part of the update, the update can be done more sparsely over blocks of size $b<B$, where $b$ is the block size and $B$ is the batch size rather than performing a $QR$-factorization for every member of the batch. In addition to the existing hyperparameters, batch size and learning rate, new hyperparameters, rank, convergence coefficient, and block size, determine the performance of the training algorithm. In traditional PCA, the starting singular vectors are randomly initialized, but in our algorithm, the previous set of singular vectors and values is used as the starting condition for the next batch. Depending on the learning rate and batch size, then, significant prior history and the rate of change of the gradient interact. 

We looked at two slightly different versions of our stochastic low rank approximation algorithm. Algorithm 1, dubbed streaming batch principal component analysis (SBPCA), uses a fixed block size $b$ and a convergence coefficient $c= 1/(i+1)$, where $i$ is the block index within a batch. To increase the stochasticity of the algorithm while minimizing the number of QR factorizations, we also investigated a variable block size version dubbed SBPCA-variable (SBPCAV) where the convergence coefficient is fixed as $c=1/2$ but the block size increases as $2^i$. The first value ($i=0$) in each block of SBPCAV will induce a large rotation in gradient space, reducing the influence of prior batch history by more strongly biasing this single randomly chosen input. SBPCA, by contrast, does not favor the ordering of inputs. Despite these differences, our investigation did not find a large difference between the performance of the two algorithms.

\begin{algorithm}
    \caption{Streaming Batch PCA (SBPCA) Update}
    \begin{algorithmic} 
        \REQUIRE $\vec\sigma$, $\hat{X}$, $\hat\Delta$ and $b$
        \STATE $\hat\Sigma=\text{diag}[\vec\sigma] $
        \FOR{$i = 1, 2, ..., B/b$}
            \STATE $\hat{y} = \hat\delta_i \hat\Delta/b$
            \STATE $\hat{z} = \hat{x}_i \hat{X}/b$
            \STATE $\hat{X} \leftarrow \frac{i\hat{X}}{i + 1} + \frac{\hat{x}_i^T \hat{y}\hat{\Sigma}^{-1}}{(i + 1)}$ $\;\;\;\;\;\;\;\;\;$ 
            \STATE $\hat\Delta \leftarrow \frac{i\hat\Delta}{i + 1} + \frac{\hat\delta_i^T \hat{z}\hat{\Sigma}^{-1}}{(i + 1) }$
            \STATE $\hat X \leftarrow {\rm QR}(\hat X)$ 
            \STATE  $\hat\Delta \leftarrow {\rm QR}(\hat\Delta)$
            \STATE  $\vec\sigma \leftarrow \frac{i \vec\sigma}{i + 1} + \sum_{\text{rows}} \frac{(\hat\delta_i \hat\Delta) \odot (\hat{x}_i \hat{X})}{(i + 1) b}$
            \STATE $\hat\Sigma=\text{diag}[\vec\sigma] $
        \ENDFOR
        \STATE Calculate $\nabla \hat\Theta = \hat{X}\hat\Sigma \hat\Delta^T$
    \end{algorithmic}
\label{alg1}
\end{algorithm}

\section{Experiments}
In this section we evaluate the two variants of the streaming batch principal component analysis (SBPCA and SBPCAV) algorithms on several standard datasets and show their results. We first briefly introduce these datasets. Then we list the detailed settings of models, followed by results and evaluation.

\subsection{Datasets}
To evaluate the streaming batch PCA approach, three image datasets of increasing size are analyzed: CIFAR-10, CIFAR-100~\cite{krizhevsky2009learning} and ImageNet~\cite{deng2009imagenet}. Fig.~\ref{Fig1} shows details of the input datasets. The CIFAR-10 dataset consists of $60\, 000$ color images of size $32\times 32$ organized into 10 classes, with $6\, 000$ images per class. The CIFAR-100 dataset is similar to CIFAR-10 dataset, with the exception that it has 100 classes containing 600 images each. The largest dataset used for this investigation is ImageNet, which consists of more than a million training images organized into $1\, 000$ classes. 
% I don't like the comma separators in the numbers above, i.e. I would just write 1,000 as 1000. -MWD
% In fact SI requires a thin space and not a comma. -MDS

\subsection{Investigated Network Structure and Hyperparameters }

\begin{figure*}[!htbp]
\centering
\includegraphics[trim=0in 13.3in 3.3in 0in,clip,width=\textwidth]{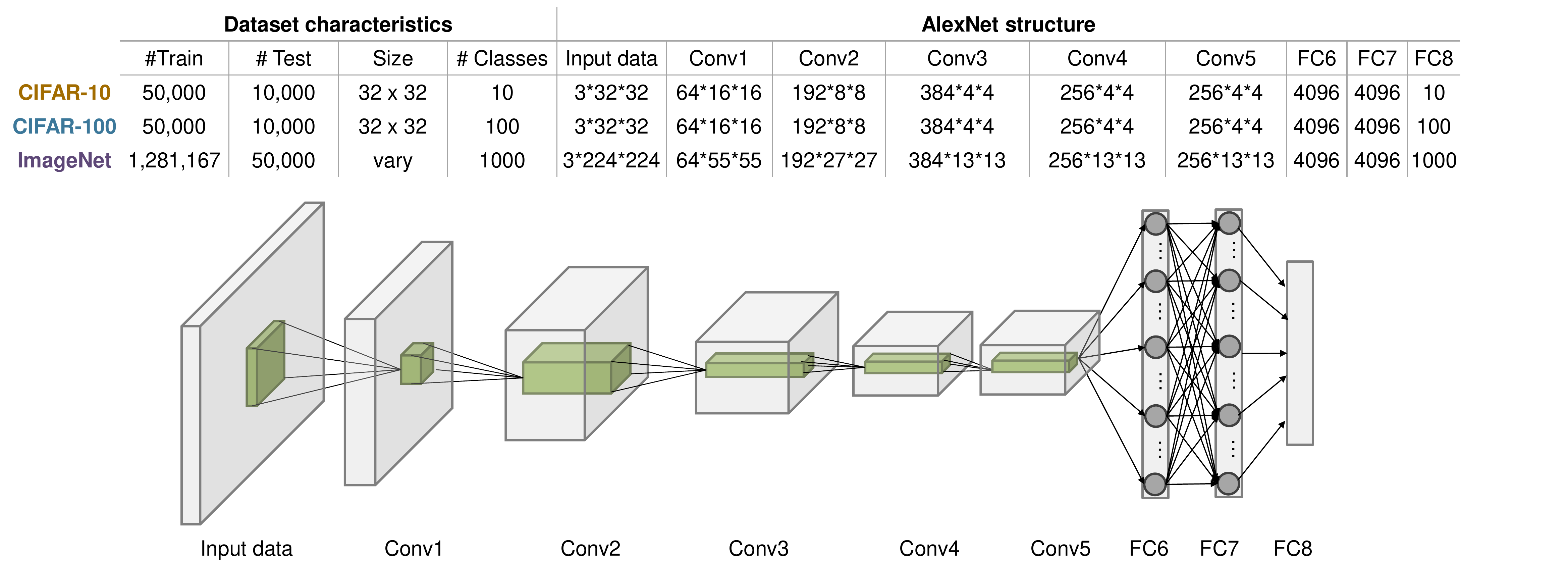}
% where an .eps filename suffix will be assumed under latex, 
% and a .pdf suffix will be assumed for pdflatex; or what has been declared
% via \DeclareGraphicsExtensions.
\caption{Dataset characteristics and modified AlexNet structure for CIFAR-10, CIFAR-100, and ImageNet}
\label{Fig1}
\end{figure*}

To investigate the algorithm, we use a well known and well understood network structure based on AlexNet~\cite{krizhevsky2012imagenet}. This network, which uses 5 convolutional layers and 3 fully connected layers, was used to achieve a high (40.9~\%) Top-5 classification rate on the ImageNet LSVRC-2010 dataset. We duplicate the structure exactly, including using the same rectifying linear unit (ReLU) activation function and four max pooling layers. The training penalty over the dataset is calculated using the traditional cross-entropy loss function for deep neural networks.  To better control for the impact of the streaming batch PCA algorithm, however, we omit certain more complex features such as regularization and data augmentation used during training in the original AlexNet implementation. To account for the smaller data set size, the fully connected layers are scaled down proportionally with image size and class size (Fig.~\ref{Fig1}). Since only the large fully connected layers have significant training data overhead, the streaming batch PCA algorithm is only implemented on these, leaving the convolutional layers to be trained with full rank batch data.

Since the streaming batch PCA algorithm is a form of gradient compression which reduces the dimensionality of the network training data, we explicitly compare it to another form of network dimensionality reduction: random dropout of neurons during training (Fig. \ref{Fig2})~\cite{dropout-reference}. While streaming batch PCA compresses the network training data in an abstract vector space, random dropout (in this case with a dropout fraction of 0.5), implements a real space reduction in both the network size and in the gradient. Note that, where streaming batch PCA and dropout are used concurrently, the activations and backpropogated errors are given no special treatment by streaming batch PCA: the dropped out neurons simply report values of 0 which are seamlessly integrated into the streaming batch PCA dataflow. 

To reduce the number of $QR$-factorizations, the block size $b$ for the streaming batch PCA approach was set to $B/4$, with $B$ the batch size, for all cases with $B$ restricted to powers of $2^n$. Where the batch size is noted for the streaming batch PCA and streaming batch PCA-variable algorithms, the batch size for the streaming batch PCA-variable approach is set to $B-1$ to accommodate the fixed convergence coefficient rule. We did not perform a search to optimize the hyperparameters, but used parameters consistent with existing practice, taking batch sizes from 128 to 2,048 and learning rates of approximately $10^{-2}$. We do notice that, at low ranks, there is significant loss of gradient information and magnitude due to the approximation which can reduce the training speed. To compensate for this, we keep the convolutional layer learning rate $\alpha_\text{conv}$ fixed at  $10^{-2}$ but investigate the impact of modifying the learning rate $\alpha_\text{fc}$ within the fully connected layers for both streaming batch PCA and MBGD.

\subsection{Results}

\subsubsection{Network structure with and without dropout}

\begin{figure}[!htbp]
\centering
\includegraphics[trim=0in 8.0in 13in 0in,clip,width=\columnwidth]{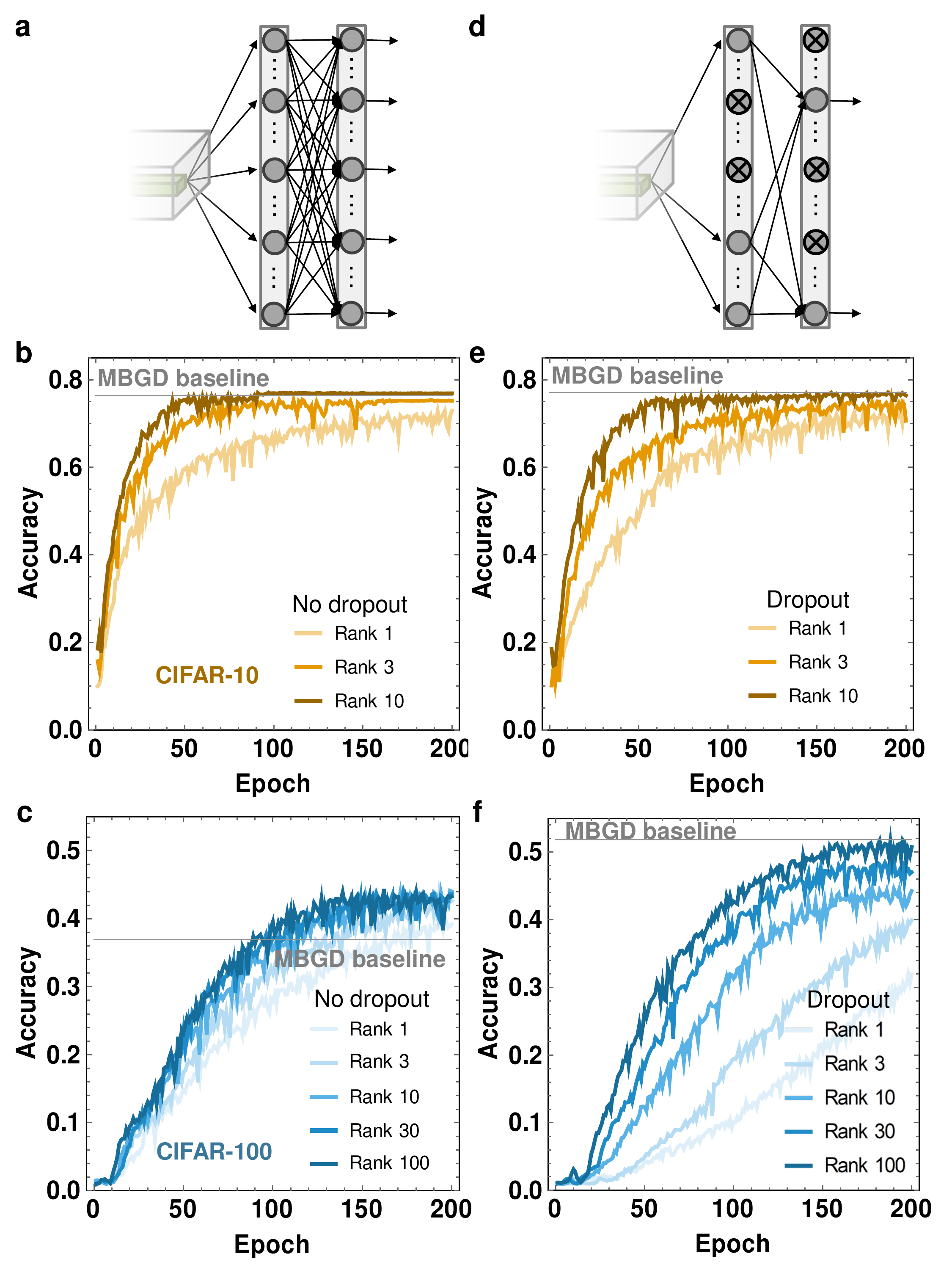}
% where an .eps filename suffix will be assumed under latex, 
% and a .pdf suffix will be assumed for pdflatex; or what has been declared
% via \DeclareGraphicsExtensions.
\caption{Network structure and rank-dependent accuracy results of SBPCA in comparison with MBGD benchmarks. (a) AlexNet structure without dropout in its fully connected layers and its accuracy results for (b) CIFAR-10 for three representative ranks (1, 3 and 10) and (c) CIFAR-100 for five representative ranks (1, 3, 10, 30, 100). (d) AlexNet structure with 50~\% dropout in its FC layers and its accuracy for (e) CIFAR-10 for ranks 1, 3, and 10 and (f) CIFAR-100 for ranks 1, 3, 10, 30, and 100. Hyperparamenters $B$ = 128, $b$ = 32, $\alpha_\text{MBGD}$ = $\alpha_\text{SBPCA}$ = 0.01.}
\label{Fig2}
\end{figure}

Fig.~\ref{Fig2} shows the accuracy for CIFAR-10 and CIFAR-100 datasets across various ranks for AlexNet with and without dropout. For CIFAR-10, the algorithm approaches comparable performance to MBGD regardless of whether or not dropout is used. For CIFAR-100, the performance for streaming batch PCA outperforms MBGD for the case with no dropout, and reaches similar performance as MBGD with dropout at higher ranks. While dropout dramatically improves the performance of networks using traditional MBGD, using dropout simultaneously with the streaming batch PCA approach does not as significantly impact the training performance.

\subsubsection{Training, overfitting, and accuracy on CIFAR-10 and CIFAR-100 }

\begin{figure}[!htbp]
\centering
\includegraphics[trim=0.2in 8.79in 13.55in 0.15in,clip,width=\columnwidth]{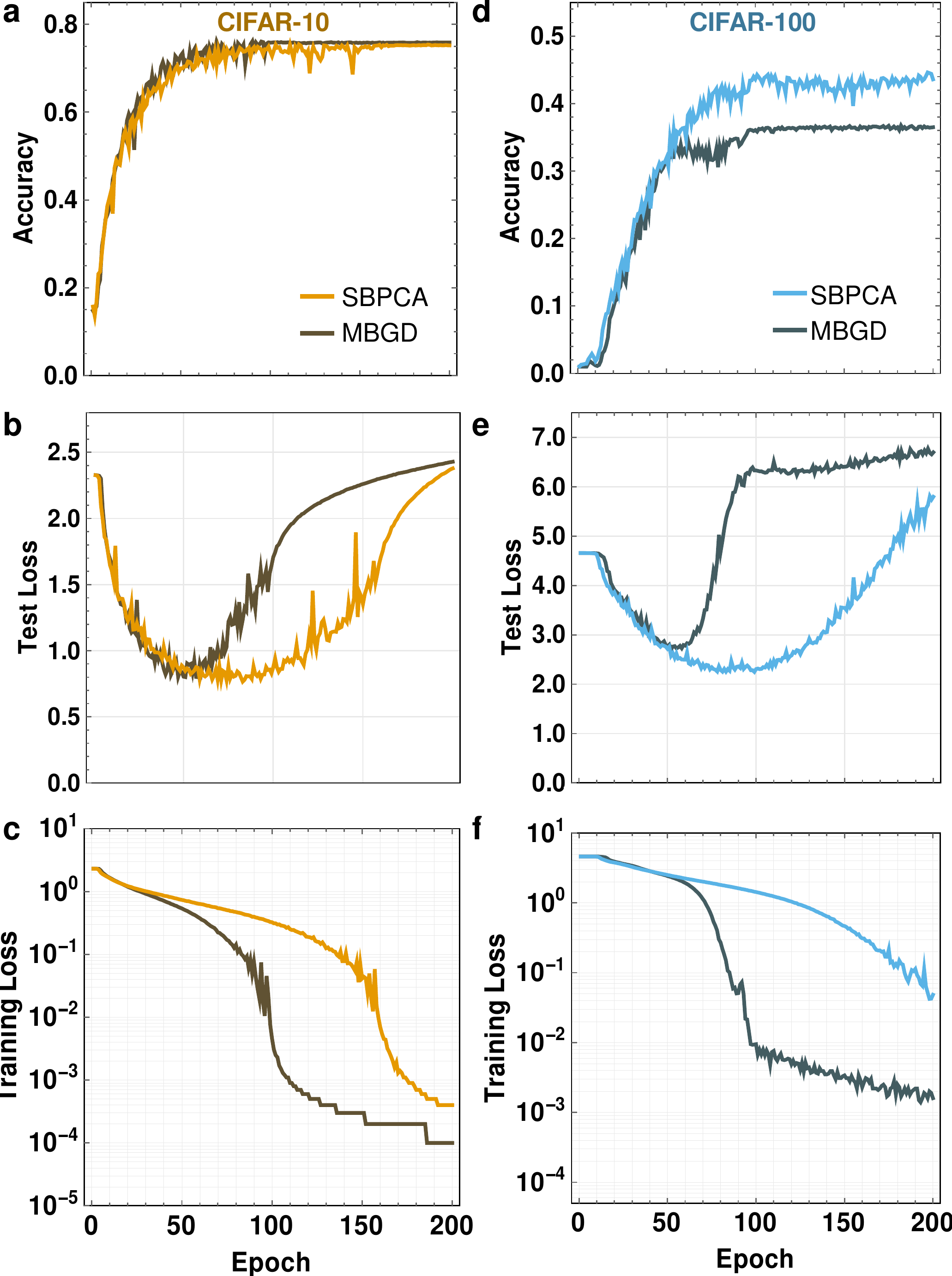}
% where an .eps filename suffix will be assumed under latex, 
% and a .pdf suffix will be assumed for pdflatex; or what has been declared
% via \DeclareGraphicsExtensions.
\caption{Network effects of proposed streaming batch PCA vs benchmark MBGD. (a) Accuracy, (b) test loss and (c) training loss for CIFAR-10, rank = 3 and (d) accuracy, (e) test loss and (f) training loss for CIFAR-100, rank = 30 respectively. The streaming batch PCA algorithm recreates the loss
minimum artifact in the AlexNet network structure also present in the MBGD. The presented results are for the case with no dropout. Hyperparameters $B$ = 128, $b$ = 32,  $\alpha_\text{MBGD}$ = $\alpha_\text{SBPCA}$ = 0.01.}
\label{Fig3}
\end{figure}

Fig. \ref{Fig3} shows the accuracy, test loss, and training set loss for CIFAR-10 and CIFAR-100 datasets for a fixed rank compared to traditional MBGD. As expected, streaming batch PCA is able to approximate the loss function gradient at a level sufficient to both reduce test-set loss and increase accuracy. Intriguingly, the performance for streaming batch PCA on CIFAR-100 significantly outperforms (by 12~\%) the accuracy from traditional MBGD at equivalent batch size and learning rate, whereas the performance difference on CIFAR-10 is negligible. % can we offer a conjecture as to why?-mwd

The key driver of the performance, however, appears to be strongly related the underlying limitations of the AlexNet structure itself rather than to the gradient descent method. While Fig.~\ref{Fig3}(c) and Fig.~\ref{Fig3}(f) show a strong tendency for the training set loss to continuously decline, the streaming batch PCA algorithm reproduces the same overfitting behavior observed in MBGD, with sharp increases in the test set loss as training continues.  In both cases, the streaming batch PCA algorithm starts out by closely tracking the MBGD algorithm before diverging as the training continues past a certain point. This tracking followed by divergence may be due to sphericalization of the loss function's eigenspectrum as the dominant sources of error are eliminated. A flatter eigenspectrum would lead to slower convergence of our power iteration methods, and would also imply that relevant gradient information is increasingly distributed beyond our $k$-rank truncated decomposition. Both factors would lead to poorer tracking of the overall gradient.  

 \subsubsection{Impact on Convergence}
\begin{figure}[!htbp]
 \centering
\includegraphics[trim=0in 8.3in 9.98in 0in,clip,width=3.54in]{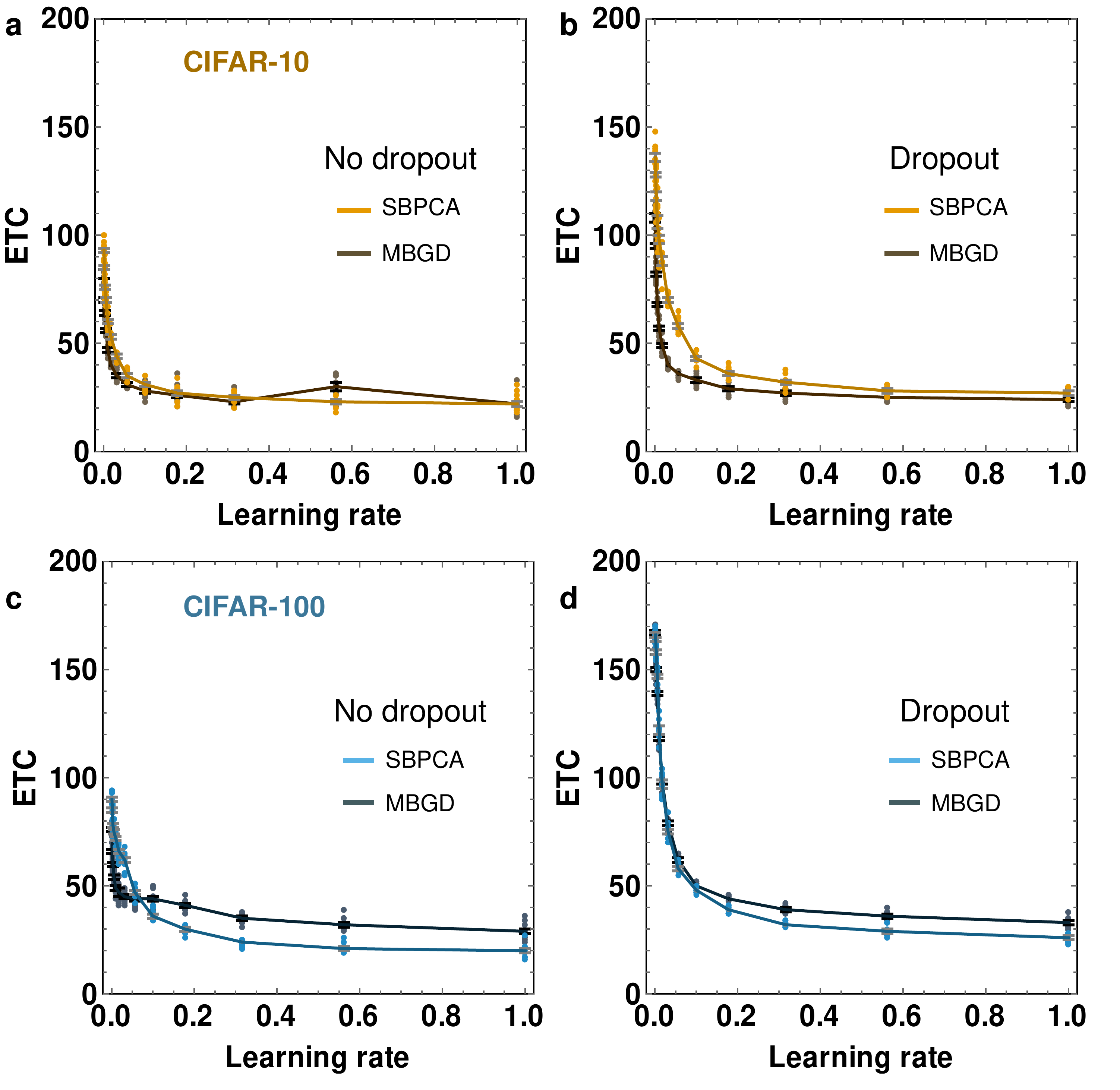}
 % where an .eps filename suffix will be assumed under latex, 
% % and a .pdf suffix will be assumed for pdflatex; or what has been declared
 % via \DeclareGraphicsExtensions.
 \caption{Epochs-to-converge (ETC) vs.~learning rate behavior for proposed streaming batch PCA and benchmark MBGD shown for (a,b) CIFAR-10 training without and with dropout  and (c,d) CIFAR-100 respectively. Ten runs were averaged to produce the plots; error bars indicate the standard deviation over trials. streaming batch PCA requires fewer epochs to converge than MBGD for lower learning rates typically used for training, and fails to converge at high learning rates. $B$ = 128, $b$ = 32.)}
 \label{Fig5}
 \end{figure}

Consistent with the notion that the streaming batch PCA training diverges from the MBGD at higher epochs, we expect that the overall time to reach final convergence should be slightly longer. Among other reasons, a truncated decomposition will have a lower overall magnitude, as implied by Eq.~\eqref{eq:frobenius-sum}. This can be partially offset relative to the convolutional layers by increasing $\alpha_\text{fc}$. In Fig.~\ref{Fig5}, we plot the epochs-to-converge (ETC) as a function of $\alpha_\text{fc}$ while leaving the convolutional layer learning rate  constant at $\alpha_\text{conv}=0.01$. Increasing the learning rate can indeed compensate for a loss in convergence rate as compared to MBGD, but MBGD can likewise improve convergence by increasing the learning rate.

\begin{figure}[!htbp]
 \centering
\includegraphics[trim=0in 17in 12.3in 0in,clip,width=3.54in]{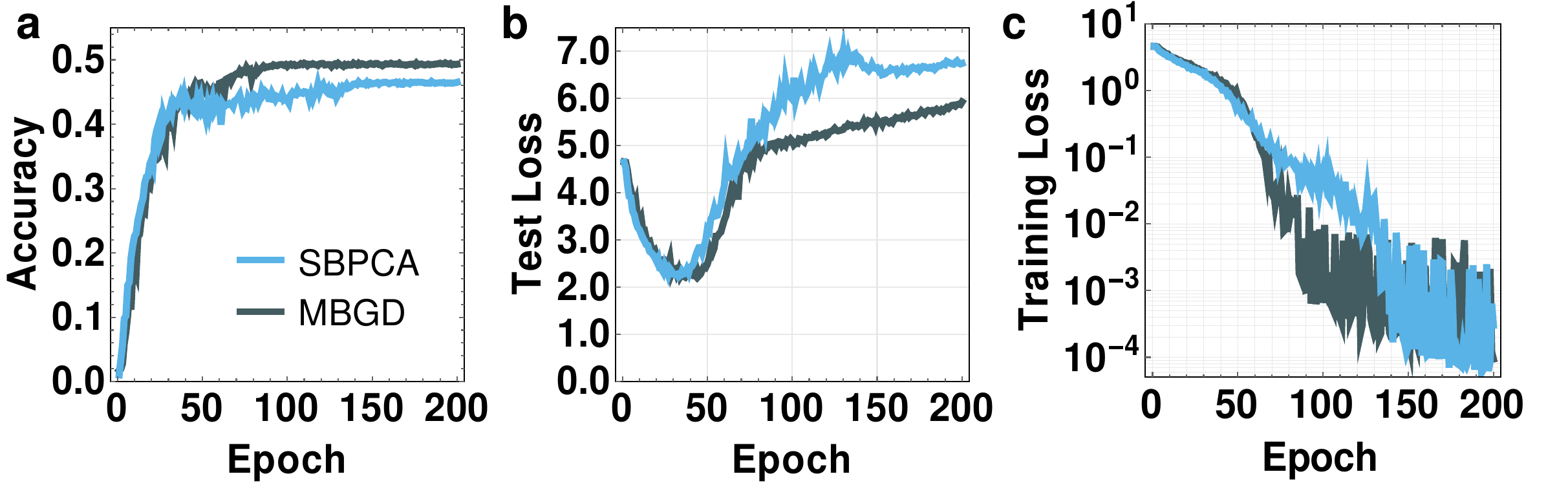}
 % where an .eps filename suffix will be assumed under latex, 
% % and a .pdf suffix will be assumed for pdflatex; or what has been declared
 % via \DeclareGraphicsExtensions.
 \caption{Network effects of the proposed streaming batch PCA vs benchmark MBGD on CIFAR-100 at higher learning rates. (a) Accuracy, (b) test loss and (c) training loss. The streaming batch PCA algorithm recreates the loss
minimum artifact in the AlexNet network structure also present in the MBGD, despite the lower performance. The presented results are with no dropout. Hyperparameters $B$ = 128, $b$ = 32, $\alpha_\text{fc}$ = 0.17, $\alpha_\text{conv}$ = 0.01 for both streaming batch PCA and MBGD.}
 \label{Fig5a}
 \end{figure}

In the case of CIFAR-100, the streaming batch PCA algorithm can converge slightly more rapidly than the MBGD algorithm, at high learning rate. Fig.~\ref{Fig5a} shows the training curves for CIFAR-100 at $\alpha_\text{fc} = 0.17$, just past the point of intersection. In this case, while streaming batch PCA exhibits superior accuracy to the MBGD at lower learning rate, MBGD is able to catch up to the accuracy of streaming batch PCA leading to a longer overall training time and explaining the crossover in the epochs-to-converge. As in Fig.~\ref{Fig3}, both the streaming batch PCA and MBGD algorithms closely track each other until the end of training and exhibit comparable overtraining artifacts. 

\subsubsection{Gradient Tracking During Training}
We further investigate the impact of our streaming batch PCA algorithm on training by extracting the rank dependence of both the accuracy and the degree of gradient tracking. The gradient tracking error can be calculated with the Frobenius norm $\Vert\nabla_{\hat\Theta}\ell -\tilde\nabla^{(k,B)}_{\hat\Theta}\ell \Vert$ where, during training, $\nabla_{\hat \Theta}\ell$ is the gradient calculated over the entire training set and $ \tilde\nabla^{(k,B)}_{\hat\Theta}\ell $ is the stochastic, $k$-rank approximation for a given batch of size $B$ calculated using the streaming batch PCA algorithm. At any given point of time during training, a lower gradient tracking error indicates superior gradient tracking. 

\begin{figure}[!htbp]
\centering
\includegraphics[trim=0in 9.18in 9.9in 0in,clip,width=3.5in]{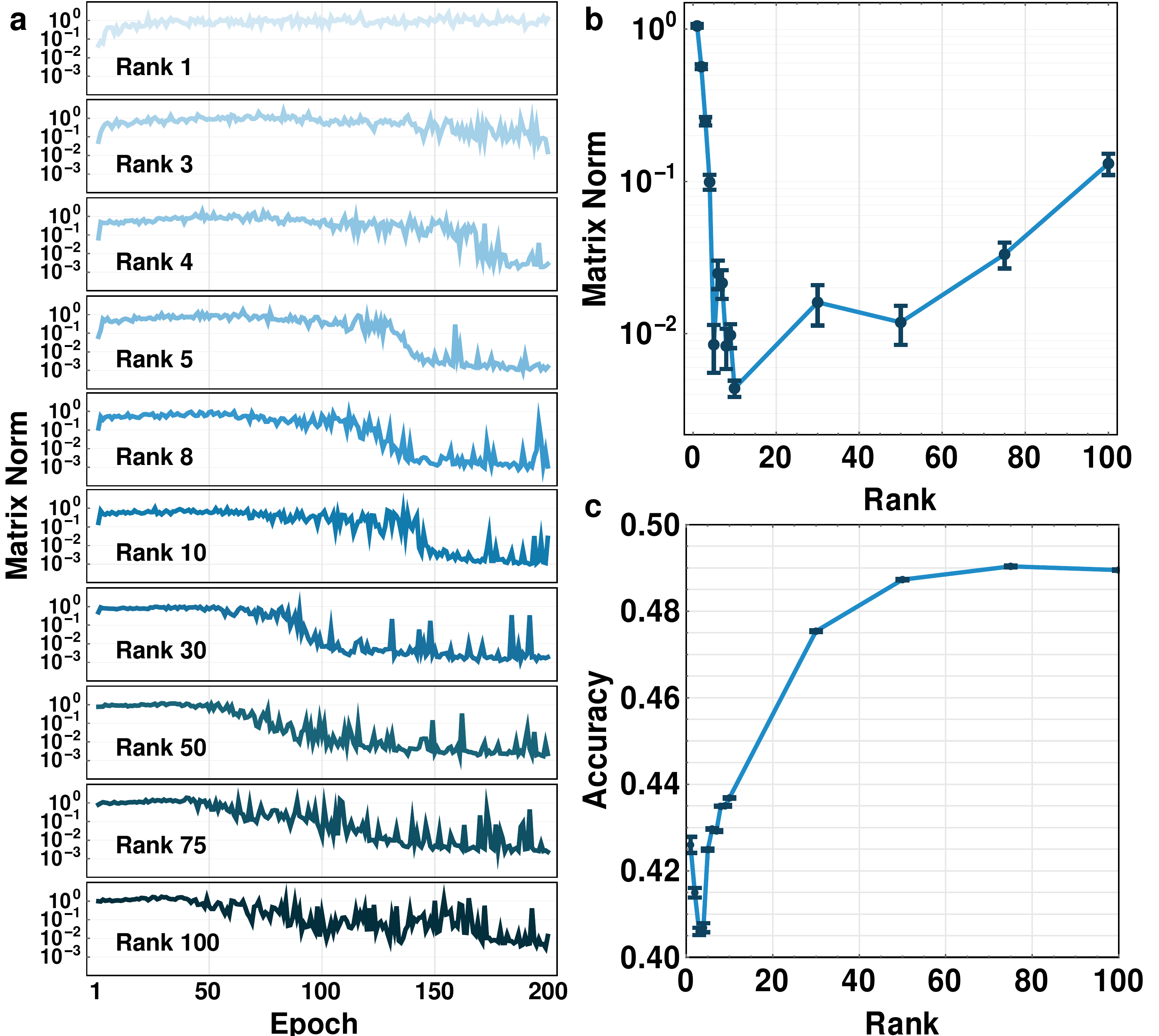}
% where an .eps filename suffix will be assumed under latex, 
% and a .pdf suffix will be assumed for pdflatex; or what has been declared
% via \DeclareGraphicsExtensions.
\caption{Matrix norm vs.~epoch vs.~rank for gradient tracking error determination. (a) The matrix norm is calculated in the second FC layer of the AlexNet during CIFAR-100 training and its dependence on epoch and rank is shown for the first 200 epochs and 10 representative ranks. (b) Gradient tracking error vs.~rank averaged over the last 50 epochs and (c) Test set accuracy vs. rank. Hyperparameters $B = 128$, $b = 32$, $\alpha_\text{fc} = 0.15$, $\alpha_\text{
conv} = 0.01$.
}
\label{Fig4}
\end{figure}

Fig.~\ref{Fig4}(a) shows the variation of the gradient tracking error as a function of rank and epoch in the last layer of our AlexNet network while training on CIFAR-100. To ensure convergence for more ranks, we set $\alpha_\text{fc} = 0.15$. In all cases, the tracking starts out initially very poorly, which we attribute to the rapid changes in the gradient as the network is updated. As the network converges, the matrix norm decreases as the network falls into a local optimum from which it cannot escape. For low ranks, the network fails to fully converge within 200 epochs and the matrix norm subsequently does not significantly decrease.

Fig.~\ref{Fig4}(b) and Fig.~\ref{Fig4}(c) show the gradient tracking error and the test-set accuracy averaged over the last 50 epochs during training. The relationship between matrix norm and network accuracy is complex and the correlation between the two is difficult to unpack. The observed behavior, however, can most likely be described as a balance between getting caught in local minima, overfitting, and random batch information. From ranks 1 to 15, the accuracy has a pronounced U-shape with increasing rank, first declining and then increasing while the matrix norm tends to decline more monotonically. This U-shape is likely due to the balance between overfitting and more efficient loss function reductions. 

Above 30 ranks, however, the matrix norm and the accuracy appear to be anti-correlated. The matrix norm likely increases as the increased number of ranks allows the matrix factorization to span a greater amount of the random batch information from batch to batch. This in turn causes the norm to be less stable from epoch to epoch, as can be seen in Fig.~\ref{Fig4}(a) where larger ranks appear noisier. This random batch information can compensate for the effects of overfitting likely caused by lower ranks over-filtering random information needed to efficiently train the network. It is likely that having few, dominant principal components which rarely change may contribute to overfitting. For this reason, having a very small number of principal components or having a very large number of principal components can lead to slightly higher accuracy as compared to having decomposition that efficiently extracts all the dominant components. 

This relationship between rank, gradient tracking error, and overall accuracy, suggests that strategies which reduce the dominant impact of the top principal components could be used to reduce overfitting of the network. Such strategies could, for instance, assign different learning rates to different principal components, minimizes the effects of overfitting by biasing different sources of network error at different levels. We do not attempt to implement any such strategy here, but they could be an interesting topic of future research.

\subsection{Comparison between Algorithms}
\subsubsection{Comparative accuracy}
\begin{figure*}[!htbp]
\centering
\includegraphics[trim=0in 9.22in 6in 0in,clip,width=5in]{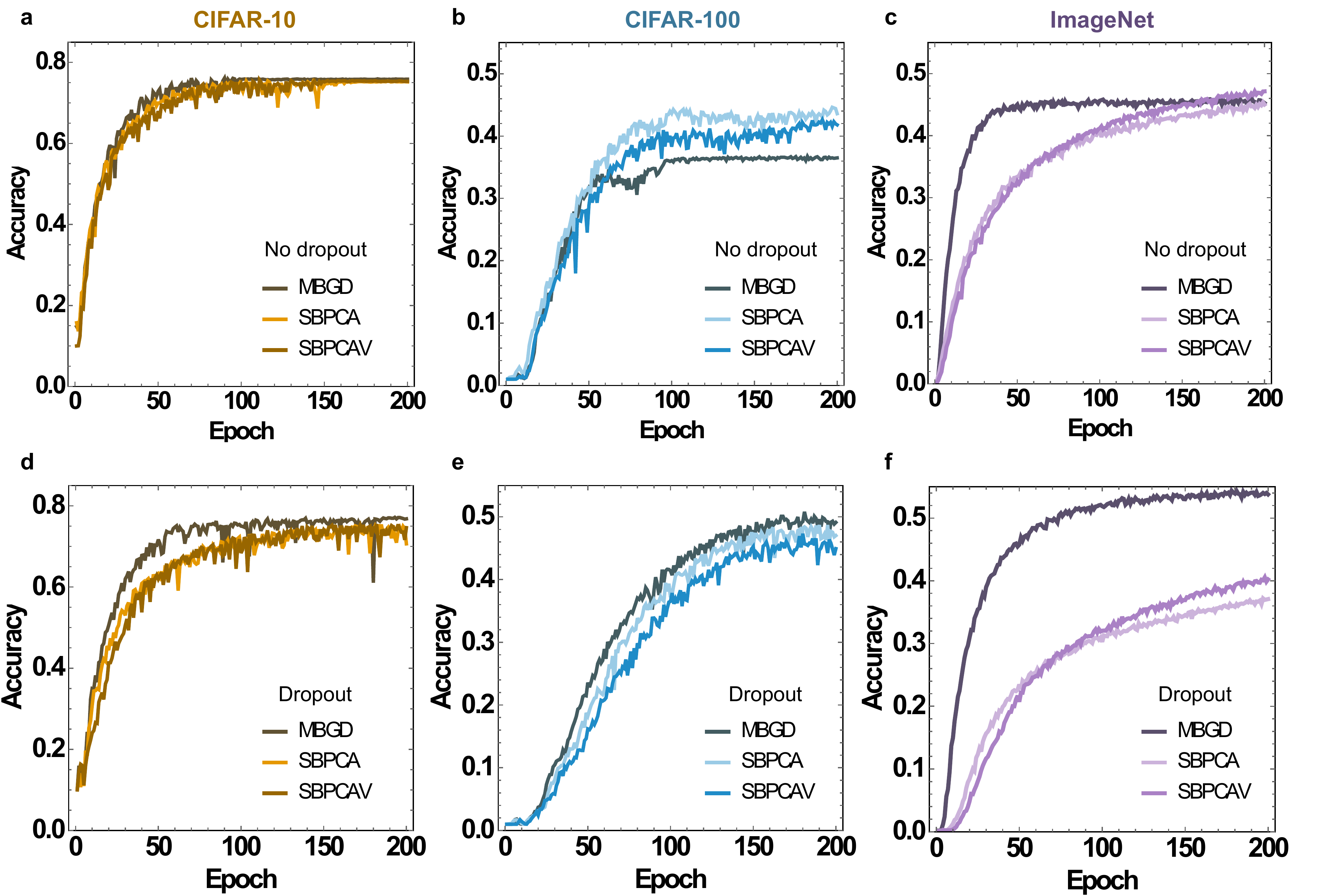}
% where an .eps filename suffix will be assumed under latex, 
% and a .pdf suffix will be assumed for pdflatex; or what has been declared
% via \DeclareGraphicsExtensions.
\caption{Streaming batch PCA and its variant streaming batch PCA-variable accuracy vs.~benchmark MBGD for three representative datasets CIFAR-10, CIFAR-100 and ImageNet (a--c) With no dropout, streaming batch PCA and streaming batch PCA-variable reach or exceed MBGD performance. (d--f) With dropout, (d) streaming batch PCA and str-variableeaming batch PCA-variable reach MBGD performance for the small database CIFAR-10; (e) streaming batch PCA outperforms streaming batch PCA-variable for the intermediate size databases CIFAR-100 reaches MBGD performance; (f) streaming batch PCA-variable performs slightly better than streaming batch PCA for the large ImageNet database, but is only capable of reaching the MBGD equivalent accuracy for long training times. CIFAR-10 and CIFAR-100 hyperparameters $B$ = 128, $b = 32$, $\alpha_\text{MBGD} = 0.01$, $\alpha_\text{SBPCA} = 0.15$. ImageNet hyperparameters $B = 256$, $b = 64$,  $\alpha_\text{MBGD} = \alpha_\text{SBPCA} = 0.01$.}
\label{Fig6}
\end{figure*}

The results from Fig.~\ref{Fig3} suggest that overfitting may be the most significant source of accuracy reduction and that it is strongly impacted by the network size. We arrive at a more detailed picture through additional comparisons between streaming batch PCA and MBGD that include a larger dataset (ImageNet) and compare against streaming batch PCA-variable and dropout. While the streaming batch approaches show equivalent [Fig.~\ref{Fig6}(a)] or superior [Fig.~\ref{Fig6}(b)] performance on small and intermediate size networks, for large scale networks [Fig.~\ref{Fig6}(c)] the improvement over MBGD is significantly reduced. The initial close tracking of MBGD vanishes in the larger network, and the streaming batch PCA and streaming batch PCA-variable approaches are only able to reach the equivalent accuracy of MBGD after four times as many epochs as MBGD.

Extending our comparison to ImageNet, however, is not entirely well controlled, as the relevant network dimension of $4000\times1000$ at the final output layer is more than ten times larger than in the CIFAR-100 case. Maintaining the same $k$ between CIFAR and ImageNet would mean that we retain a considerably smaller fraction of the eigenspectrum in ImageNet; a fair comparison would hold constant the ratio between $k$ and the layer size. Unfortunately, achieving comparable ratios would mean keeping several hundred ranks in ImageNet, which is computationally impractical due to the tenfold baseline increase in QR factorization time arising from the larger network size. Compounding this with an additional tenfold increase in the number of ranks, while managing proportional demands on memory, would would have exceeded the computational resources reasonably available to us in the absence of an application specific integrated circuit (ASIC) optimized for performing PCA on the incoming data. Though the requisite number of calculations may be lower than calculating the full batch matrix, the vector matrix operations required to perform MBGD can be effectively pipelined in the existing GPU architecture. 

While increasing the number of ranks might have allowed the streaming batch PCA approach to match or exceed the accuracy of MBGD, a computationally efficient approach to increasing the network accuracy, given our resources, is dropout, which allows the network to avoid overfitting. While the smallest network, CIFAR-10, has almost no change in performance when using dropout, the MBGD performance on both CIFAR-10 and CIFAR-100 dramatically increases. No such equivalent increase in accuracy is seen when combining dropout with the streaming batch PCA or streaming batch PCA-variable algorithms, suggesting that the prior history information stored within the streaming batch PCA gradient approximations mitigates the advantages gained from randomly varying the network topology to train the network. Our choice to combine streaming batch PCA with dropout without excluding the dropped-out neurons from the update could be suboptimal; alternative approaches to combining the algorithms could fare better.

\subsubsection{Memory and Computational Efficiency}
\begin{figure}[!htbp]
\centering
\includegraphics[trim=0.1in 14.24in 11.6in 0in,clip,width=\columnwidth]{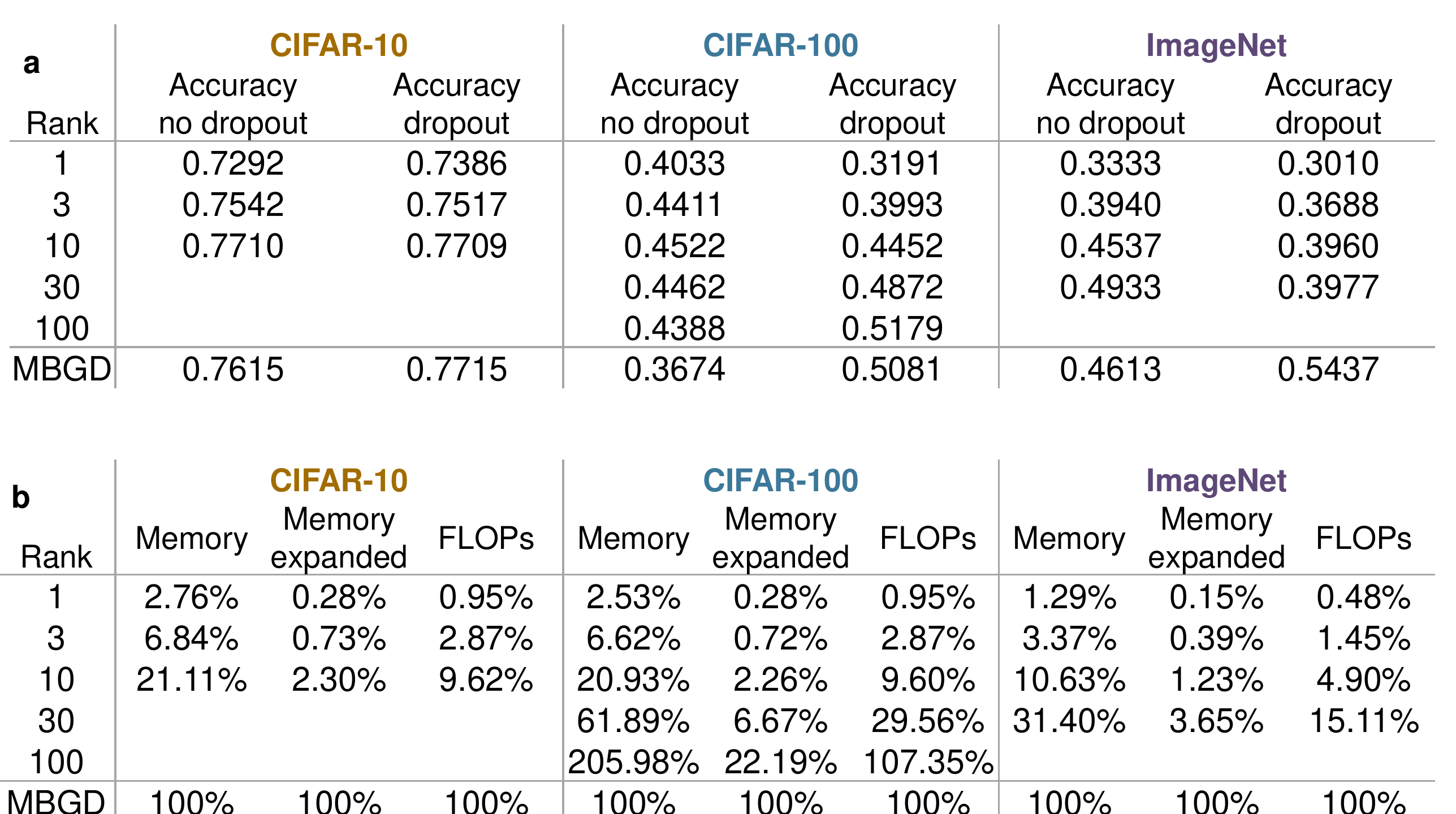}
\caption{Accuracy vs. memory cost. (a) Accuracy comparison for streaming batch PCA vs. MBGD on for CIFAR-10; CIFAR-100 and ImageNet. (b) Memory estimates for streaming batch PCA vs.~MBGD on for CIFAR-10; CIFAR-100 and ImageNet. streaming batch PCA has highly efficient memory utilization (only 1~\% to 15~\% of MBGD) with little loss in accuracy. No dropout structure. CIFAR-10 and CIFAR-100 hyperparameters $B = 128$, $b = 32$; ImageNet hyperparameters $B = 256$, $b = 64$. For all three datasets  $\alpha_\text{MBGD} = \alpha_\text{SBPCA} = 0.01$.}
\label{Fig7}
\end{figure}

While the accuracy differences between the streaming batch PCA and MBGD can be small [Table~\ref{Fig7}(a)], the streaming batch PCA approach has significant memory and computational advantages over MBGD, significantly reducing the calculation overhead [Table~\ref{Fig7}(b)]. Without dedicated hardware, however, these advantages may not necessarily translate into wall clock or energy savings.

The reduction in memory overhead is the most significant, especially when compared to explicit calculation of the batch gradient. For an $m\times n$ matrix calculated over a batch of size $B$, the pre-calculated memory overhead scales with $B(m+ n)$ whereas the fully calculated gradient scales with $m\times n$, the overall network size. In contrast, the streaming batch PCA memory overhead requires storage of $k$ memory locations for the singular values and $k (m+n)$ storage for the left and right singular vectors. The runtime operation of the streaming batch PCA operation requires additional memory overhead. The memory storage needed to store the updates to the gradients over a streaming block of size $b$ is an additional  $k(m+n)$ storage units. More importantly, the $QR$-factorization algorithm itself requires additional memory which could theoretically be as low as $m$ for the case of the memory re-using Householder requirement~\cite{schreiber1989storage}. In practice, however, the requirement that the $QR$-factorization algorithm be unique and explicitly calculated requires that the $R$ matrix also be explicitly calculated to guarantee that its diagonal entries are non-increasing.  The traditional Householder algorithm, which produces the $R$ matrix, requires a memory overhead of $k^2 +\max(m,n)(k+1)$, with the $\max$ function requiring that the additional $QR$-factorization memory being able to fit the larger $QR$ matrix. For small $k$, the memory reduction would be approximately $(3k+1)/B$ for the non-explicitly calculated batch update or a quadratically smaller $[2k(m+n)+\max(m,n)(k+1)+k^2]/(mn)$ for the expanded matrix. 

Computationally, there are also significant reductions in the overall number of floating point operations (FLOPs) needed for the calculations. Whereas MBGD requires $2Bmn$ FLOPs to produce the batch update, the streaming batch PCA approach requires $4Bk(m+n)$ calculations to produce the updates to the singular vectors and an additional ${B}/{b}$ $QR$-factorizations to reorthonormalize the vectors. Other more negligible steps, with complexity scaling as $Bk$, are needed for the convergence coefficient and singular values. For sufficiently small $k$, the $QR$-factorizations will asymptotically require $O(2k^2(m+n))$ FLOPs to orthonormalize both matrices~\cite{golub2012matrix}. After the batch, expanding the matrix requires $2kmn$ calculations to  update the neural network layer. In the asymptotic limit of large $m$ and $n$ and small $k$, the computational overhead could be reduced by a factor of $k/B$.

Based on these estimates, Fig.~\ref{Fig7} summarizes the associated memory and computational reductions of calculating the network updates from the activations and errors versus the network performance of the streaming batch PCA algorithm \emph{vis a vis} MGBD as a function of network size. For fair comparison, the accuracy numbers are derived from the experiments employing dropout, as dropout is an efficient means of improving network performance with low computational overhead. In general, larger networks experience more efficient utilization of the required memory overhead at the expense of a significant reduction in the accuracy. This reduction could be offset by going to a larger number of ranks, but this is both computationally unfavorable and practically intractable due to the inefficient calculations of the $QR$-factorization in current hardware. Dedicated hardware such as an ASIC or field programmable gate array could realize these theoretical gains. 

Notwithstanding the existing reductions in the computational and memory overhead, additional modifications to the algorithm could yield significant additional advantages. Combining the streaming batch PCA approach with, for example, implicit updates or by including the streaming batch PCA vectors in the subsequent batch calculations and update can be useful to further avoid the large $m\times n$ scale calculations. These calculations are required to update the main matrix and are the most computationally intensive part of the update calculation. Additionally, extending this approach to low rank approximations of other methods, such as momentum and AdaGrad, may yield further increases in efficiency. The approach investigated here has so far only been the simplest possible use of the streaming batch PCA algorithm within the context of training neural networks, and additional investigation promise to yield further gains.

\section{Conclusion}
 In this paper, we propose streaming batch principal component analysis as an algorithmic approach to training hardware neural networks. By producing a stochastic, $k$-rank approximation of the gradient using bi-iterative stochastic power iterations, we show both theoretically and experimentally that it is possible to reduce the memory and computational overhead of training neural networks using real data sets. Though the lower rank algorithm can have somewhat lower performance than its full rank counter part, the primary sources of the performance loss may be related more strongly to the weaknesses of the overall network architecture (such as from overfitting), rather than a fundamental flaw with the training approach itself. Additional theoretical investigation, especially by extending the algorithm to the network convolutional layers considering the implications of network topology, are still needed. Nevertheless, our results suggest that future machine learning algorithms and hardware can exploit low-rank approaches to compressing the training data to realize real improvements in the energy and memory requirements of network training.

% or


\section*{Acknowledgment}

The authors would like to thank Jabez McClelland, Nikolai Zhitnev, Advait Madhavan, and Mark Anders for useful scientific discussions and the \textit{enki} team for computational support. This work has been supported by the DARPA / ONR grant No. N00014-20-1-2031 and GWU University Facilitating Fund.

\ifCLASSOPTIONcaptionsoff
  \newpage
\fi



%\bibliographystyle{IEEEtran}

\bibliography{ref.bib}

\begin{thebibliography}{10}
\providecommand{\url}[1]{#1}
\csname url@samestyle\endcsname
\providecommand{\newblock}{\relax}
\providecommand{\bibinfo}[2]{#2}
\providecommand{\BIBentrySTDinterwordspacing}{\spaceskip=0pt\relax}
\providecommand{\BIBentryALTinterwordstretchfactor}{4}
\providecommand{\BIBentryALTinterwordspacing}{\spaceskip=\fontdimen2\font plus
\BIBentryALTinterwordstretchfactor\fontdimen3\font minus
  \fontdimen4\font\relax}
\providecommand{\BIBforeignlanguage}[2]{{%
\expandafter\ifx\csname l@#1\endcsname\relax
\typeout{** WARNING: IEEEtran.bst: No hyphenation pattern has been}%
\typeout{** loaded for the language `#1'. Using the pattern for}%
\typeout{** the default language instead.}%
\else
\language=\csname l@#1\endcsname
\fi
#2}}
\providecommand{\BIBdecl}{\relax}
\BIBdecl

\bibitem{shallue2019measuring}
C.~J. Shallue, J.~Lee, J.~Antognini, J.~Sohl-Dickstein, R.~Frostig, and G.~E.
  Dahl, ``Measuring the effects of data parallelism on neural network
  training,'' \emph{Journal of Machine Learning Research}, vol.~20, pp. 1--49,
  2019.

\bibitem{mayer2020scalable}
R.~Mayer and H.-A. Jacobsen, ``Scalable deep learning on distributed
  infrastructures: Challenges, techniques, and tools,'' \emph{ACM Computing
  Surveys (CSUR)}, vol.~53, no.~1, pp. 1--37, 2020.

\bibitem{li2014communication}
M.~Li, D.~G. Andersen, A.~J. Smola, and K.~Yu, ``Communication efficient
  distributed machine learning with the parameter server,'' in \emph{Advances
  in Neural Information Processing Systems}, 2014, pp. 19--27.

\bibitem{wen2017terngrad}
W.~Wen, C.~Xu, F.~Yan, C.~Wu, Y.~Wang, Y.~Chen, and H.~Li, ``Terngrad: Ternary
  gradients to reduce communication in distributed deep learning,'' in
  \emph{Advances in neural information processing systems}, 2017, pp.
  1509--1519.

\bibitem{kataeva2019towards}
I.~Kataeva, S.~Ohtsuka, H.~Nili, H.~Kim, Y.~Isobe, K.~Yako, and D.~Strukov,
  ``Towards the development of analog neuromorphic chip prototype with 2.4 m
  integrated memristors,'' in \emph{2019 IEEE International Symposium on
  Circuits and Systems (ISCAS)}.\hskip 1em plus 0.5em minus 0.4em\relax IEEE,
  2019, pp. 1--5.

\bibitem{burraccelerating}
G.~W. Burr, S.~Ambrogio, P.~Narayanan, H.~Tsai, C.~Mackin, and A.~Chen,
  ``Accelerating deep neural networks with analog memory devices.''

\bibitem{adam20163}
G.~C. Adam, B.~D. Hoskins, M.~Prezioso, F.~Merrikh-Bayat, B.~Chakrabarti, and
  D.~B. Strukov, ``3-d memristor crossbars for analog and neuromorphic
  computing applications,'' \emph{IEEE Transactions on Electron Devices},
  vol.~64, no.~1, pp. 312--318, 2016.

\bibitem{adam2018challenges}
G.~C. Adam, A.~Khiat, and T.~Prodromakis, ``Challenges hindering memristive
  neuromorphic hardware from going mainstream,'' \emph{Nature communications},
  vol.~9, no.~1, pp. 1--4, 2018.

\bibitem{chakrabarti2017multiply}
B.~Chakrabarti, M.~A. Lastras-Monta{\~n}o, G.~Adam, M.~Prezioso, B.~Hoskins,
  M.~Payvand, A.~Madhavan, A.~Ghofrani, L.~Theogarajan, K.-T. Cheng
  \emph{et~al.}, ``A multiply-add engine with monolithically integrated 3d
  memristor crossbar/cmos hybrid circuit,'' \emph{Scientific reports}, vol.~7,
  p. 42429, 2017.

\bibitem{prezioso2015training}
M.~Prezioso, F.~Merrikh-Bayat, B.~Hoskins, G.~C. Adam, K.~K. Likharev, and
  D.~B. Strukov, ``Training and operation of an integrated neuromorphic network
  based on metal-oxide memristors,'' \emph{Nature}, vol. 521, no. 7550, pp.
  61--64, 2015.

\bibitem{hoskins2019streaming}
B.~D. Hoskins, M.~W. Daniels, S.~Huang, A.~Madhavan, G.~C. Adam, N.~Zhitenev,
  J.~J. McClelland, and M.~Stiles, ``Streaming batch eigenupdates for hardware
  neural networks,'' \emph{Frontiers in Neuroscience}, vol.~13, p. 793, 2019.

\bibitem{yang1995extension}
B.~Yang, ``An extension of the pastd algorithm to both rank and subspace
  tracking,'' \emph{IEEE Signal Processing Letters}, vol.~2, no.~9, pp.
  179--182, 1995.

\bibitem{mitliagkas2013memory}
I.~Mitliagkas, C.~Caramanis, and P.~Jain, ``Memory limited, streaming pca,'' in
  \emph{Advances in neural information processing systems}, 2013, pp.
  2886--2894.

\bibitem{wang2019characterizing}
M.~Wang, C.~Meng, G.~Long, C.~Wu, J.~Yang, W.~Lin, and Y.~Jia, ``Characterizing
  deep learning training workloads on alibaba-pai,'' \emph{arXiv preprint
  arXiv:1910.05930}, 2019.

\bibitem{lin2017deep}
Y.~Lin, S.~Han, H.~Mao, Y.~Wang, and W.~J. Dally, ``Deep gradient compression:
  Reducing the communication bandwidth for distributed training,'' \emph{arXiv
  preprint arXiv:1712.01887}, 2017.

\bibitem{jouppi2017datacenter}
N.~P. Jouppi, C.~Young, N.~Patil, D.~Patterson, G.~Agrawal, R.~Bajwa, S.~Bates,
  S.~Bhatia, N.~Boden, A.~Borchers \emph{et~al.}, ``In-datacenter performance
  analysis of a tensor processing unit,'' in \emph{Proceedings of the 44th
  Annual International Symposium on Computer Architecture}, 2017, pp. 1--12.

\bibitem{suri2013bio}
M.~Suri, D.~Querlioz, O.~Bichler, G.~Palma, E.~Vianello, D.~Vuillaume,
  C.~Gamrat, and B.~DeSalvo, ``Bio-inspired stochastic computing using binary
  cbram synapses,'' \emph{IEEE Transactions on Electron Devices}, vol.~60,
  no.~7, pp. 2402--2409, 2013.

\bibitem{bishop2019monolithic}
M.~D. Bishop, H.-S.~P. Wong, S.~Mitra, and M.~M. Shulaker, ``Monolithic 3-d
  integration,'' \emph{IEEE Micro}, vol.~39, no.~6, pp. 16--27, 2019.

\bibitem{li2018capacitor}
Y.~Li, S.~Kim, X.~Sun, P.~Solomon, T.~Gokmen, H.~Tsai, S.~Koswatta, Z.~Ren,
  R.~Mo, C.~Yeh \emph{et~al.}, ``Capacitor-based cross-point array for analog
  neural network with record symmetry and linearity,'' in \emph{2018 IEEE
  Symposium on VLSI Technology}.\hskip 1em plus 0.5em minus 0.4em\relax IEEE,
  2018, pp. 25--26.

\bibitem{oja1985stochastic}
E.~Oja and J.~Karhunen, ``On stochastic approximation of the eigenvectors and
  eigenvalues of the expectation of a random matrix,'' \emph{Journal of
  mathematical analysis and applications}, vol. 106, no.~1, pp. 69--84, 1985.

\bibitem{liu1989limited}
D.~C. Liu and J.~Nocedal, ``On the limited memory bfgs method for large scale
  optimization,'' \emph{Mathematical programming}, vol.~45, no. 1-3, pp.
  503--528, 1989.

\bibitem{guan2012online}
N.~Guan, D.~Tao, Z.~Luo, and B.~Yuan, ``Online nonnegative matrix factorization
  with robust stochastic approximation,'' \emph{IEEE Transactions on Neural
  Networks and Learning Systems}, vol.~23, no.~7, pp. 1087--1099, 2012.

\bibitem{oja1982simplified}
E.~Oja, ``Simplified neuron model as a principal component analyzer,''
  \emph{Journal of mathematical biology}, vol.~15, no.~3, pp. 267--273, 1982.

\bibitem{oja1992principal}
------, ``Principal components, minor components, and linear neural networks,''
  \emph{Neural networks}, vol.~5, no.~6, pp. 927--935, 1992.

\bibitem{yang2018history}
P.~Yang, C.-J. Hsieh, and J.-L. Wang, ``History pca: A new algorithm for
  streaming pca,'' \emph{arXiv preprint arXiv:1802.05447}, 2018.

\bibitem{hardt2014noisy}
M.~Hardt and E.~Price, ``The noisy power method: A meta algorithm with
  applications,'' in \emph{Advances in Neural Information Processing Systems},
  2014, pp. 2861--2869.

\bibitem{li2016rivalry}
C.-L. Li, H.-T. Lin, and C.-J. Lu, ``Rivalry of two families of algorithms for
  memory-restricted streaming pca,'' in \emph{Artificial Intelligence and
  Statistics}, 2016, pp. 473--481.

\bibitem{allen2017first}
Z.~Allen-Zhu and Y.~Li, ``First efficient convergence for streaming k-pca: a
  global, gap-free, and near-optimal rate,'' in \emph{2017 IEEE 58th Annual
  Symposium on Foundations of Computer Science (FOCS)}.\hskip 1em plus 0.5em
  minus 0.4em\relax IEEE, 2017, pp. 487--492.

\bibitem{balcan2016improved}
M.-F. Balcan, S.~S. Du, Y.~Wang, and A.~W. Yu, ``An improved gap-dependency
  analysis of the noisy power method,'' in \emph{Conference on Learning
  Theory}, 2016, pp. 284--309.

\bibitem{strobach1997bi}
P.~Strobach, ``Bi-iteration svd subspace tracking algorithms,'' \emph{IEEE
  Transactions on signal processing}, vol.~45, no.~5, pp. 1222--1240, 1997.

\bibitem{krizhevsky2009learning}
A.~Krizhevsky, G.~Hinton \emph{et~al.}, ``Learning multiple layers of features
  from tiny images,'' Citeseer, Tech. Rep., 2009.

\bibitem{deng2009imagenet}
J.~Deng, W.~Dong, R.~Socher, L.-J. Li, K.~Li, and L.~Fei-Fei, ``Imagenet: A
  large-scale hierarchical image database,'' in \emph{2009 IEEE conference on
  computer vision and pattern recognition}.\hskip 1em plus 0.5em minus
  0.4em\relax Ieee, 2009, pp. 248--255.

\bibitem{krizhevsky2012imagenet}
A.~Krizhevsky, I.~Sutskever, and G.~E. Hinton, ``Imagenet classification with
  deep convolutional neural networks,'' in \emph{Advances in neural information
  processing systems}, 2012, pp. 1097--1105.

\bibitem{dropout-reference}
N.~Srivastava, G.~Hinton, A.~Krizhevsky, I.~Sutskever, and R.~Salakhutdinov,
  ``Dropout: a simple way to prevent neural networks from overfitting,''
  \emph{The journal of machine learning research}, vol.~15, no.~1, pp.
  1929--1958, 2014.

\bibitem{schreiber1989storage}
R.~Schreiber and C.~Van~Loan, ``A storage-efficient wy representation for
  products of householder transformations,'' \emph{SIAM Journal on Scientific
  and Statistical Computing}, vol.~10, no.~1, pp. 53--57, 1989.

\bibitem{golub2012matrix}
G.~H. Golub and C.~F. Van~Loan, \emph{Matrix computations}.\hskip 1em plus
  0.5em minus 0.4em\relax JHU press, 2012, vol.~3.

\end{thebibliography}
\bibliographystyle{IEEEtran}


\vspace{-12 mm}
\begin{IEEEbiography}[{\includegraphics[width=1in,height=1.25in,clip,keepaspectratio]{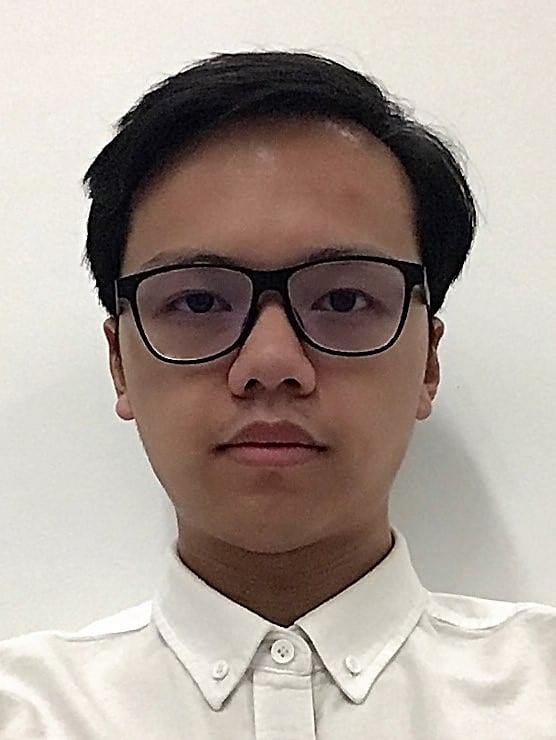}}]{Siyuan ``Yin'' Huang}
is a M.S. student in Computer Science at George Washington University and has a B.S. degree in Software Engineering from Wuhan University of Technology received in 2018. He works in the Adaptive Devices and Microsystems group at GWU since Fall 2018 on algorithms to minimize the impact of device non-idealities on memristive neural networks.
\end{IEEEbiography}%
\vspace{-10 mm}
\begin{IEEEbiography}[{\includegraphics[width=1in,height=1.25in,clip,keepaspectratio]{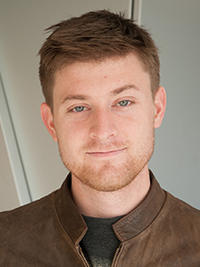}}]{Brian D. Hoskins}is a research physicist in the Alternative Computing Group at National Institute of Standards and Technology (NIST) Gaithersburg, USA. He received both a B.S. and an M.S. in Materials Science and Engineering from Carnegie Mellon University in 2010 and 2011 respectively and a Ph.D. in Materials from the University of California, Santa Barbara in 2016. For his doctoral research, he developed and characterized resistive switching devices for use in neuromorphic networks. Brian is working on CMOS integration of resistive switches for the development and characterization of intermediate scale neuromorphic networks.\end{IEEEbiography}%
\vspace{-10 mm}
\begin{IEEEbiography}[{\includegraphics[width=1in,height=1.25in,clip,keepaspectratio]{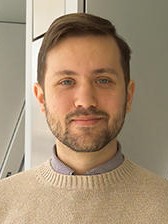}}]{Matthew W. Daniels} is a National Research Council (NRC) Postdoctoral Fellow in the Alternative Computing Group at the National Institute for Standards and Technology (NIST). He received a B.S. in Physics with a minor in Mathematics from Clemson University, and M.S. and Ph.D. degrees in Physics from Carnegie Mellon University. His dissertation work was on antiferromagnetic magnon physics. His postdoctoral work touches multiple topics in neuromorphic computing, with a focus on discovering and utilizing new information encodings that bridge the gap between traditional CMOS engineering and the energy efficient, native computing capabilities of magnetic and memristive nanotechnology. 
\end{IEEEbiography}%
\vspace{-10 mm}
\begin{IEEEbiography}[{\includegraphics[width=1in,height=1.25in,clip,keepaspectratio]{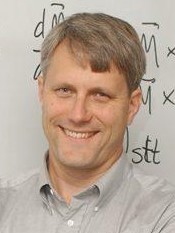}}]{Mark D. Stiles}is a NIST Fellow in the Alternative Computing Group at the National Institute of Standards and Technology. He received a M.S./B.S. in Physics from Yale University, and M.S. and Ph.D. degrees in Physics from Cornell University. Following postdoctoral research at AT\&T Bell Laboratories, he joined the research staff at NIST. Mark's research, published in over 145 papers, has focused on the development of a variety of theoretical methods for predicting the properties of magnetic nanostructures and has recently shifted to neuromorphic computing. He has helped organize numerous conferences and has served the American Physical Society as the Chair of the Topical Group on Magnetism and on its Executive Committees as well as that of the Division of Condensed Matter Physics. He has also served Physical Review Letters as a Divisional Associate Editor, served on the Editorial Board of Physical Review Applied, and is currently an Associate Editor for Reviews of Modern Physics. Mark is a Fellow of the American Physical Society, and has been awarded the Silver Medal from the Department of Commerce.%
\end{IEEEbiography}
\vspace{-10 mm}
\begin{IEEEbiography}[{\includegraphics[width=1in,height=1.25in,clip,keepaspectratio]{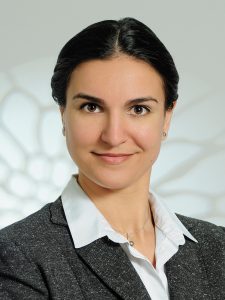}}]{Gina C. Adam}is an assistant professor in the Electrical and Computer Engineering department at George Washington University, USA. She leads the Adaptive Devices and Microsystems group, that works on the development of novel hardware foundations using emerging nanodevices to enable new ways of computing. She received her Ph.D. in electrical and computer engineering from the University of California, Santa Barbara in 2015 and was a research scientist at the Romanian National Institute for Research and Development in Microtechnologies and a visiting scholar at École Polytechnique Fédérale de Lausanne before joining GWU. She was the recipient of an International Fulbright Science and Technology award in 2010, a Mirzayan fellowship at the National Academy of Engineering in 2012 and of a H2020 Marie Sklodowska-Curie grant from the European Commission in 2016-2018.%
\end{IEEEbiography}
\vfill


%\vfill


\end{document}